\documentclass[conference]{IEEEtran}
\IEEEoverridecommandlockouts
\usepackage{cite}
\usepackage{amsmath,amssymb,amsfonts}
\usepackage{algorithmic}
\usepackage{graphicx}
\usepackage{textcomp}
\usepackage{xcolor}
\usepackage{tikz}
\usetikzlibrary{positioning,arrows.meta,fit,calc}
\usepackage{pgfplots}
\pgfplotsset{compat=1.16}
\usepackage{booktabs}
\usepackage{array}
\usepackage[font=footnotesize,labelsep=colon,justification=justified,singlelinecheck=false,skip=3pt]{caption}
\begin{document}
\title{Benchmarking Vision-Language Models for Urdu Nastaliq Handwriting Recognition}
\author{
\IEEEauthorblockN{Haseeb Yaqoob}
\IEEEauthorblockA{\textit{Department of Computer Science}\\
\textit{NED University of Engineering \& Technology}\\
Karachi, Pakistan\\
haseeb.yaqoob201@gmail.com}
\and
\IEEEauthorblockN{Fatima Azfar}
\IEEEauthorblockA{\textit{Department of Computer Science}\\
\textit{NED University of Engineering \& Technology}\\
Karachi, Pakistan\\
Fatimaazfar08@gmail.com}
\and
\IEEEauthorblockN{Emaan Arif Khan}
\IEEEauthorblockA{\textit{Department of Computer Science}\\
\textit{NED University of Engineering \& Technology}\\
Karachi, Pakistan\\
eamyarifkhan@gmail.com}
}
\maketitle
\begin{abstract}
Handwritten Urdu is hard to recognize because the cursive Nastaliq script has no fixed baseline and no clear character boundaries. We benchmark six systems on the Urdu Nastaliq Handwritten Dataset (UNHD): Tesseract (default configuration), a script-specific recognizer trained on printed Urdu (UTRNet-Large), two open-source vision-language models (VLMs), and two commercial VLMs. All six systems are evaluated without fine-tuning on 2,465 single-line images (Part A), and the four VLMs receive the same fixed prompt (decoding uses temperature 0 where configurable, except for Claude Sonnet 5, whose decoding cannot be configured). Both commercial VLMs reach lower error than the printed-Urdu recognizer and the open-source VLMs. We do not compare against handwriting-trained recognizers, which would likely do better in-domain; we characterize zero-shot behavior, not state of the art. We also describe three distinct failure modes among the ranked systems. In one of them, a commercial model explicitly declines to transcribe a few very hard-to-read images instead of guessing. In a second study (Part B), VLMs correct the output of a separate OCR system using both the source image and the noisy text. Once separated from each model's own reading ability, the benefit is small on Tesseract's output (0.7\% to 7.4\%) and larger (17.9\% and 21.9\%) only for the two commercial correctors given UTRNet's text. It appears to depend on how legible the supplied text already is, and a weaker corrector can make a more legible text worse.
\end{abstract}
\section{Introduction}
\label{sec:intro}
Urdu is written in the Nastaliq script. Nastaliq is cursive and flows diagonally, and the shape of a character depends on its position in the word and on its neighbors. These properties make it hard to split a line into characters, which makes handwritten Urdu text recognition (HTR) difficult. Existing work trains script-specific recognizers on the few public datasets available~\cite{unhd,zia2022,ganai2023}.

General-purpose vision-language models (VLMs) are trained on broad multilingual and multimodal data, and we apply them without any Urdu-specific fine-tuning of our own.

We study two research questions:
\begin{itemize}
    \item \textbf{RQ1:} Can zero-shot, general-purpose VLMs match or exceed Tesseract and a recognizer trained for printed Urdu on handwritten Urdu Nastaliq? Which accuracy and failure characteristics distinguish them?
    \item \textbf{RQ2:} Can VLMs improve a separate OCR system's output as a zero-shot correction layer, given both the source image and that system's noisy transcription, rather than replacing that system outright?
\end{itemize}

Although multimodal language models have been benchmarked for handwritten text recognition across several languages, their performance on Urdu Nastaliq remains less explored. We evaluate general-purpose vision-language models on the Urdu Nastaliq Handwritten Dataset (UNHD) under a controlled zero-shot protocol and investigate whether these models can improve existing OCR outputs when provided with both the source image and the initial transcription.

We answer RQ1 in \textbf{Part A}, a controlled benchmark of six systems on the Urdu Nastaliq Handwritten Dataset (UNHD) (Sections~\ref{sec:method} and~\ref{sec:parta}). A ranking of systems does not answer RQ2. RQ2 is a separate question that matters for deployment, and we study it in \textbf{Part B} (Section~\ref{sec:partb}).

\section{Related Work}
\label{sec:related}
Handwritten Urdu recognition has long been limited by scarce public datasets~\cite{unhd}. NUST-UHWR was introduced with a CNN-RNN model with n-gram language modeling~\cite{zia2022}, and a Long-term Recurrent Convolutional Network was evaluated directly on UNHD~\cite{ganai2023}, which also noted that UNHD's 10,000 nominal lines contain only a small fraction of semantically unique ones (Section~\ref{sec:dataset}).

More recent work uses transformers: convolution-transformer hybrids and unified printed-and-handwritten models~\cite{riaz2022,maqsood2023}, an efficient transformer reporting state-of-the-art CER~\cite{hamza2024}, a paragraph-level ViT-decoder with a language model~\cite{dey2025bharatocr}, and a CRNN baseline on the historical UKHD dataset~\cite{basharat2026}. All are trained for this task; we do not fine-tune the VLMs we evaluate for Urdu HTR.

The OCR ability of general-purpose VLMs has been benchmarked across scripts and tasks~\cite{ocrbench}, mostly on Latin-script and high-resource text. Two studies compare zero-shot multimodal LLMs with specialized HTR models on non-English or historical scripts, with mixed results: proprietary models excel on modern English handwriting but show no consistent advantage on historical non-English text~\cite{crosilla2025}, and zero-shot Gemini is only comparable to a minimally trained TrOCR baseline on historical correspondence~\cite{li2024chr}. Neither covers Urdu Nastaliq, which we address here.

\section{Dataset}
\label{sec:dataset}
We use the Urdu Nastaliq Handwritten Dataset (UNHD)~\cite{unhd}. The raw corpus has 7,610 scanned single-line images and 7,396 ground-truth files. After resolving orphaned pairs and removing 37 groups of exact duplicate images (re-checking labels where duplicates disagreed), 7,319 valid pairs remain, spanning 849 unique sentence templates, consistent with earlier work showing that UNHD's nominal line count overstates its diversity~\cite{ganai2023}.

Our evaluation set takes 2--3 distinct writers per sentence. To avoid counting one physical writer twice, we manually grouped writer IDs that seemed to belong to the same person into clusters. When a sentence had too few distinct writers, we drew an extra sample from a cluster already used for it (a ``same-cluster fallback writer''; 16 sentences). Twelve sentences with fewer than two available writers even with this fallback were excluded. The final set has 2,465 image/ground-truth pairs across 837 sentences and 336 distinct writers. Evaluating on all UNHD samples without limiting repetition can be dominated by a few heavily repeated templates, so our protocol limits the influence of repeated sentences and writers (the writer clustering was manual and approximate).

\section{Methodology}
\label{sec:method}
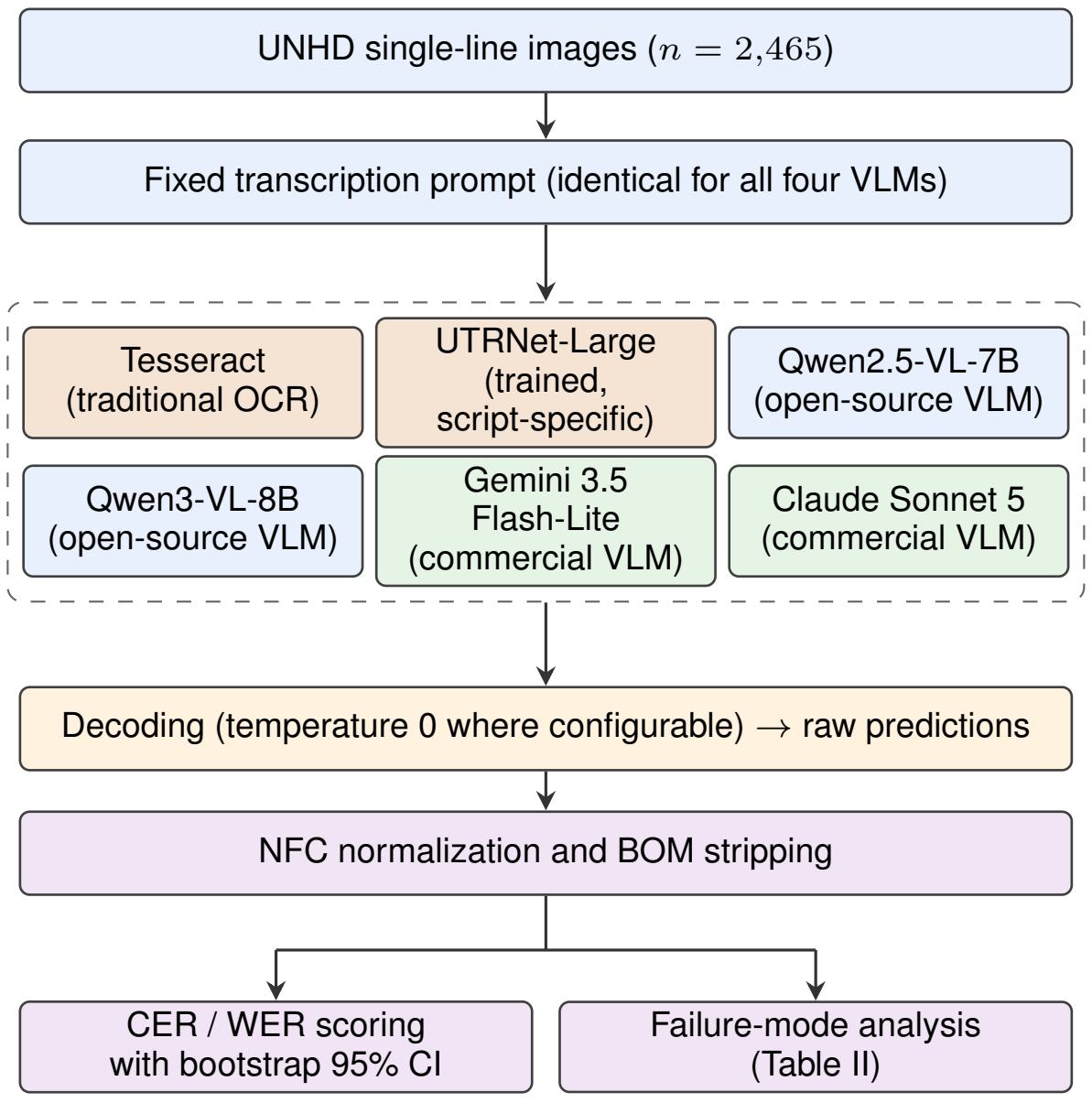
\begin{figure}[t]
\centering
\definecolor{figblue}{HTML}{E8F0FE}
\definecolor{figpeach}{HTML}{F5E4D3}
\definecolor{figgreen}{HTML}{E6F4E6}
\definecolor{figcream}{HTML}{FFF3DF}
\definecolor{figlav}{HTML}{F0E4F5}
\begin{tikzpicture}[
    font=\sffamily\scriptsize, align=center,
    nb/.style={draw=black!75, rounded corners=2pt, line width=0.5pt, inner sep=2pt, minimum height=0.6cm},
    sb/.style={draw=black!75, rounded corners=2pt, line width=0.5pt, inner sep=2pt, minimum width=2.45cm, minimum height=0.8cm},
    arr/.style={-{Stealth[length=4pt,width=4pt]}, line width=0.6pt, black!80},
    ln/.style={line width=0.6pt, black!80}]
\node[nb, fill=figblue, minimum width=7.6cm] (img) at (0,0) {UNHD single-line images ($n = 2{,}465$)};
\node[nb, fill=figblue, minimum width=7.6cm] (prompt) at (0,-0.95) {Fixed transcription prompt (identical for all four VLMs)};
\node[sb, fill=figpeach] (s1) at (-2.55,-2.4) {Tesseract\\(traditional OCR)};
\node[sb, fill=figpeach] (s2) at (0,-2.4) {UTRNet-Large\\(trained,\\script-specific)};
\node[sb, fill=figblue]  (s3) at (2.55,-2.4) {Qwen2.5-VL-7B\\(open-source VLM)};
\node[sb, fill=figblue]  (s4) at (-2.55,-3.4) {Qwen3-VL-8B\\(open-source VLM)};
\node[sb, fill=figgreen] (s5) at (0,-3.4) {Gemini 3.5\\Flash-Lite\\(commercial VLM)};
\node[sb, fill=figgreen] (s6) at (2.55,-3.4) {Claude Sonnet 5\\(commercial VLM)};
\node[draw=black!60, dashed, rounded corners=3pt, inner sep=3pt, fit=(s1)(s2)(s3)(s4)(s5)(s6)] (grp) {};
\node[nb, fill=figcream, minimum width=7.6cm] (raw) at (0,-4.9) {Decoding (temperature 0 where configurable) $\rightarrow$ raw predictions};
\node[nb, fill=figlav, minimum width=7.6cm] (norm) at (0,-5.8) {NFC normalization and BOM stripping};
\node[nb, fill=figlav, minimum width=3.7cm] (cer) at (-1.95,-7.2) {CER / WER scoring\\with bootstrap 95\% CI};
\node[nb, fill=figlav, minimum width=3.7cm] (qual) at (1.95,-7.2) {Failure-mode analysis\\(Table~\ref{tab:failure_modes})};
\draw[arr] (img) -- (prompt);
\draw[arr] (prompt) -- (grp.north);
\draw[arr] (grp.south) -- (raw);
\draw[arr] (raw) -- (norm);
\draw[ln] (norm.south) -- (0,-6.5);
\draw[ln] (-1.95,-6.5) -- (1.95,-6.5);
\draw[arr] (-1.95,-6.5) -- (cer.north);
\draw[arr] (1.95,-6.5) -- (qual.north);
\end{tikzpicture}
\caption{Part A pipeline.}
\label{fig:pipeline}
\end{figure}

Fig.~\ref{fig:pipeline} shows the Part A pipeline. The four VLMs receive the same fixed prompt, while Tesseract and UTRNet-Large run with their default or official settings. Decoding is greedy or temperature 0 (or a fixed default) for every system except Claude Sonnet 5, which does not expose decoding parameters (see below). Raw predictions are normalized, scored with CER and WER, and analyzed for failure modes (Table~\ref{tab:failure_modes}).

\subsection{Model Roster}
The systems span an untrained general-purpose OCR engine, a script-specific supervised recognizer, and two tiers of zero-shot VLMs, which tests how far generic multimodal pretraining can replace Urdu-specific supervision. Six systems were evaluated end to end and ranked (Table~\ref{tab:parta_results}):
\begin{itemize}
    \item \textbf{Tesseract}~\cite{tesseract} (version \texttt{4.1.1}): a widely used open-source OCR engine, run with the \texttt{urd} language pack and default settings (\texttt{pytesseract.image\_to\_string(img, lang="urd")}), with no image preprocessing and no page-segmentation override. Tesseract's default full-page mode was therefore applied to single-line images, which likely understates its best performance.
    \item \textbf{UTRNet-Large}~\cite{utrnet}: an HRNet-based recognizer trained for Urdu script, run with its official inference configuration. UTRNet was proposed and evaluated for \textit{printed} Urdu. Here we use it across domains as a handwriting baseline.
    \item \textbf{Qwen2.5-VL-7B-Instruct}~\cite{qwen25vl} and \textbf{Qwen3-VL-8B-Instruct}~\cite{qwen3vl}: open-source VLMs, run with vLLM with 4-bit bitsandbytes quantization in fp16 on a single NVIDIA Tesla T4 GPU.
    \item \textbf{Gemini 3.5 Flash-Lite}~\cite{gemini} and \textbf{Claude Sonnet 5}~\cite{claude}: commercial VLMs, run through their official APIs with metered, pay-per-use batch inference.
\end{itemize}
We also evaluated two further open-source VLMs, MiniCPM-V-4.5~\cite{minicpmv45} and InternVL3.5-8B~\cite{internvl35}, both as 4-bit quantized checkpoints. We excluded them from the ranking (Section~\ref{sec:negative}).

\subsection{Fixed Prompt and Decoding Policy}
We want a zero-shot capability comparison, not a per-model prompt-tuning exercise. Every VLM therefore received the same instruction, delivered as the system message (or the API's system-instruction field) with the image in the user turn:
\begin{quote}
\textit{``This image contains a single line of handwritten Urdu text in the Nastaliq script. Transcribe exactly what is written, character for character, including any diacritics. Output ONLY the transcribed Urdu text, with no translation, no explanation, and no extra commentary.''}
\end{quote}

\noindent The open-source models used greedy, deterministic generation (temperature 0), a repetition penalty of 1.05, and a 128-token cap. Some models fell into repetition loops. The mild penalty was the only repetition control we used under vLLM; a stronger setting (penalty 1.3 with a no-repeat-3-gram constraint) was tried in an early \texttt{transformers} pilot and discarded because it induced hallucinated commentary. The penalty was applied identically to all open-source models and was not per-model tuning.

Gemini 3.5 Flash-Lite ran at temperature 0 with \texttt{max\_output\_tokens} $=128$ and its default thinking setting; outputs averaged 19.8 tokens (maximum 40), so none reached the cap. Claude Sonnet 5 ran through the Anthropic Batch API at the high-resolution image tier (2576\,px maximum edge) with \texttt{max\_tokens} $=1024$. Its model generation does not allow temperature, top\_p, or top\_k to be set (setting them returns an error), so its decoding relies on built-in behavior.

Extended thinking is on by default for Sonnet 5, and an explicit \texttt{thinking: \{"type": "disabled"\}} request did not always take effect in preliminary tests. The 1024-token cap is therefore a shared budget for thinking and output. The first pass used \texttt{max\_tokens} $=128$, and 294 responses were truncated because thinking consumed the whole budget; those rows were resubmitted with the larger cap and all recovered. This matters for the empty/refusal rate in Table~\ref{tab:parta_results} and the unusable-output rate in Section~\ref{sec:catastrophic}. The API model IDs were \texttt{gemini-3.5-flash-lite} and \texttt{claude-sonnet-5}.

\textbf{Implementation details.} The open-source models ran under vLLM 0.25.1 on a single Kaggle Tesla T4 with each model's default image processor, a 4096-token maximum length, and batches of 16 images. UTRNet-Large used the authors' released checkpoint with the official preprocessing (grayscale, height 32, left-right flip). Each system was run once; API outputs at temperature 0 need not be identical across runs, and Claude Sonnet 5's run-to-run variance is not quantified because its decoding cannot be fixed. Gemini rows that failed on quota errors (320) were resubmitted unchanged. All reported numbers come from the common scoring script in Section~\ref{sec:metrics}.

\subsection{Evaluation Metrics}
\label{sec:metrics}
Our primary metrics are diacritic-sensitive CER and WER, each the edit distance divided by the reference length, computed per image and averaged over images from each system's raw prediction file. We recomputed all aggregates independently from the raw per-image CSVs, checking row counts, unique image identifiers, and the absence of unexplained errors.

The scoring script decodes ground truth with \texttt{utf-8-sig}, which strips the byte-order mark at the start of about 51\% of raw ground-truth files, and applies Unicode NFC normalization to predictions and ground truth before alignment. Skipping either step would add a small systematic error to every system that comes from encoding artifacts.

To check whether differences are more than sampling noise, we report 95\% confidence intervals from a 10,000-resample paired percentile bootstrap over images~\cite{efron1993}. Images are resampled with replacement, jointly across the two quantities being compared, so per-image pairing is preserved. Each interval runs from the 2.5th to the 97.5th percentile of the statistic (mean CER, or CER reduction as a percentage of the baseline).

\section{Part A: Benchmarking Results}
\label{sec:parta}
\subsection{Aggregate Accuracy}
Table~\ref{tab:parta_results} reports mean CER, WER, and empty or refusal rates for the six ranked systems.

Both commercial VLMs outperformed the printed-Urdu recognizer and the open-source VLMs. Gemini 3.5 Flash-Lite had the lowest CER and Claude Sonnet 5 was next. Every adjacent-rank difference was statistically significant under paired image-level bootstrap resampling, so the ranking is unlikely to come from image-sampling variability. The smallest gap was between Claude Sonnet 5 and UTRNet-Large (paired difference $-0.043$, 95\% CI $[-0.051,-0.033]$). Section~\ref{sec:discussion} reports a sentence-level check of this comparison.

Claude Sonnet 5's CER leaves out the five images it refused (Section~\ref{sec:failuremodes}). If those images are scored as maximal error (CER $=1.0$), its CER is 0.2671. This does not change the ranking.

Both commercial VLMs beat UTRNet-Large, a recognizer trained on printed Urdu, without any Urdu-specific fine-tuning on our part. This suggests that broad multimodal pretraining can offset much of the need for task-specific training, although we cannot rule out that UNHD appeared in their pretraining data (Section~\ref{sec:discussion}).

\begin{table*}[t]
\caption{Part A results ($N = 2{,}465$). CER and WER are diacritic-sensitive. CIs are 10,000-resample paired bootstrap intervals.}
\label{tab:parta_results}
\centering
\footnotesize
\begin{tabular*}{\textwidth}{@{}l@{\extracolsep{\fill}}lccc c@{}}
\toprule
\textbf{System} & \textbf{Category} & \textbf{CER} & \textbf{95\% CI (CER)} & \textbf{WER} & \textbf{Empty / refusal} \\
\midrule
Tesseract & Traditional OCR & 0.7215 & [0.7145, 0.7286] & 1.1793 & 5.1\% \\
Qwen2.5-VL-7B & Open-source VLM & 0.6464 & [0.6343, 0.6592] & 0.9495 & 0\% \\
Qwen3-VL-8B & Open-source VLM & 0.4861 & [0.4737, 0.4990] & 0.8380 & 0\% \\
UTRNet-Large & Trained (script-specific) & 0.3084 & [0.3017, 0.3153] & 0.7827 & 0\% \\
Claude Sonnet 5 & Commercial VLM & 0.2656 & [0.2550, 0.2769] & 0.5225 & 0.20\% \\
Gemini 3.5 Flash-Lite & Commercial VLM & \textbf{0.1718} & [0.1633, 0.1805] & \textbf{0.3772} & 0\% \\
\bottomrule
\end{tabular*}
\end{table*}

\subsection{Qualitative Failure Modes}
\label{sec:failuremodes}
Beyond the ranking, the six systems show three distinct failure modes (Table~\ref{tab:failure_modes}). Gemini 3.5 Flash-Lite showed none of them on this evaluation set.

Claude Sonnet 5's behavior stands out. On 5 of 2,465 images (0.20\%), it explicitly declined to transcribe instead of guessing, each time with an English explanation that the handwriting was too hard to read. Each was retried once and none recovered, so the refusals appear repeatable, though we retried only once. Table~\ref{tab:parta_results} counts them as refusals, not empty output. We inspected the five images by hand: they were unusually hard to read even for a fluent Urdu reader, which suggests real uncertainty and not an instruction-following failure.

Repetition-loop rates in Table~\ref{tab:failure_modes} count outputs in which a phrase of one to four words repeats three or more times in a row. They are Part A rates, and are not directly comparable to the Part B rates in Section~\ref{sec:catastrophic}, which use a separate detector on corrector outputs.

\begin{table}[t]
\caption{Failure modes and their rates in Part A.}
\label{tab:failure_modes}
\centering
\footnotesize
\begin{tabular}{@{}>{\raggedright\arraybackslash}p{0.72in}>{\raggedright\arraybackslash}p{0.55in}>{\raggedright\arraybackslash}p{1.05in}>{\raggedright\arraybackslash}p{0.6in}@{}}
\toprule
\textbf{Failure mode} & \textbf{Seen in} & \textbf{Description} & \textbf{Rate} \\
\midrule
Empty output & Tesseract & Likely full-page mode on single lines (single-line mode untested) & 5.1\% \\
Repetition loop & Qwen2.5-VL, Qwen3-VL & Phrase repeated until the generation cap & 1.66\% / 0.81\% \\
Refusal on hard images & Claude Sonnet 5 & Explicit decline to transcribe & 0.20\% \\
None observed & Gemini 3.5 Flash-Lite & No empty or refused output & 0\% \\
\bottomrule
\end{tabular}
\end{table}

\subsection{Negative Results: MiniCPM-V-4.5 and InternVL3.5-8B}
\label{sec:negative}
We evaluated two further open-source VLMs and left them out of Table~\ref{tab:parta_results}. InternVL3.5-8B (third-party AWQ 4-bit checkpoint) obtained a mean CER of 1.0633 (WER 1.3206) on all 2,465 images, worse than an empty prediction; a partial rerun with the official checkpoint under bitsandbytes ($n=152$) gave 1.0779. MiniCPM-V-4.5 (official int4 checkpoint) was tested only on a small pilot because of its failure pattern: a handful of images under vLLM (CER about 1.5--3.2, often with unrequested English commentary) and three through its native \texttt{transformers} interface (CER 0.71--0.78), so we report no full-set CER. InternVL3.5-8B produces fluent, grammatical Urdu unrelated to the handwritten source, a failure mode we call \textit{confident hallucination}. The MiniCPM-V-4.5 pilot is too small and unstable to say whether it shows the same pattern. The InternVL result likely reflects a capability limit on this task, although quantization and integration effects cannot be fully ruled out.

\section{Part B: Zero-Shot Correction}
\label{sec:partb}
Text-only LLM correctors trained on synthetic noise can greatly reduce CER on conventional OCR output~\cite{guan2024}, but a study on historical English and Finnish found the benefit neither universal nor guaranteed~\cite{kanerva2025noFL}. Multimodal correction, which also gives the corrector the source image, outperformed text-only correction on historical German city directories and reached sub-1\% CER without fine-tuning~\cite{greif2025}. We apply that design to zero-shot Urdu Nastaliq and find a real but more conditional benefit (Section~\ref{sec:crosscorrector}).

\subsection{Experimental Setup}
Part B asks whether VLMs can improve a separate OCR system's output when given both the source image and that system's noisy transcription. All correctors use the image-plus-text setting: the source image and the noisy OCR text, with no other context. (A text-only ablation, reported in Section~\ref{sec:discussion}, withholds the image.) All correctors, on both input sources, receive the same correction instruction as the system prompt:
\begin{quote}
\textit{``You are correcting the output of a low-quality OCR system that transcribed a single line of handwritten Urdu text in the Nastaliq script. The OCR system's guess is often wrong -- garbled, incomplete, or entirely incorrect. Your job is to produce the correct transcription. Output ONLY the corrected Urdu text -- no explanation, no commentary.''}
\end{quote}

\noindent The user turn contains the source image and the line ``OCR guess: $\langle$text$\rangle$'' (image first for the Qwen models and Claude Sonnet 5, text first for Gemini). The open-source correctors used temperature 0, a repetition penalty of 1.05, and a 128-token cap, as in Part A, and Gemini ran at temperature 0 with the same cap. Claude Sonnet 5 ran through the Anthropic Batch API under the same model ID as in Part A, with the instruction as the system prompt, the original PNG sent base64-encoded without client-side resizing, and \texttt{max\_tokens} $=1024$ from the first pass to avoid the Part A truncation problem. As in Part A, no temperature or thinking setting was passed, so its decoding relies on built-in behavior and run-to-run variance is not quantified. Rows that returned an API error, an empty output, or an English non-transcription response were resubmitted once in a targeted retry and merged; rows still empty after the retry are the unusable outputs counted in Section~\ref{sec:catastrophic}.

We test four correctors: Qwen2.5-VL-7B-Instruct, Qwen3-VL-8B-Instruct, Gemini 3.5 Flash-Lite, and Claude Sonnet 5. Each corrects two inputs. The first is Tesseract's Part A predictions, the weakest baseline. The second is UTRNet-Large's Part A predictions, a much more accurate baseline. The second input tests whether the effect, and its direction, still hold when the text prior is already considerably more legible. We keep each corrector's own image-only CER from Part A, on the same images, as a reference point.

\subsection{What Does ``Percent Reduction'' Actually Measure?}
\label{sec:reframing}
\begin{figure*}[t]
\centering
\definecolor{bgray}{HTML}{C8C8C8}
\definecolor{bblue}{HTML}{4C7EA8}
\begin{tikzpicture}
\begin{axis}[
    ybar=8pt, bar width=15pt,
    width=0.97\textwidth, height=5cm,
    xmin=0.45, xmax=8.55, ymin=-17, ymax=90,
    ytick={0,20,40,60,80},
    yticklabel={\pgfmathprintnumber[fixed,precision=0,assume math mode=true]{\tick}},
    extra y ticks={0}, extra y tick labels={},
    extra y tick style={grid=major, grid style={solid,black!80,line width=0.4pt}},
    xtick={1,...,8},
    xticklabels={
      {\shortstack{Qwen2.5-\\VL-7B\\(Tesseract input)}},
      {\shortstack{Qwen2.5-\\VL-7B\\(UTRNet input)}},
      {\shortstack{Qwen3-\\VL-8B\\(Tesseract input)}},
      {\shortstack{Qwen3-\\VL-8B\\(UTRNet input)}},
      {\shortstack{Gemini 3.5\\Flash-Lite\\(Tesseract input)}},
      {\shortstack{Gemini 3.5\\Flash-Lite\\(UTRNet input)}},
      {\shortstack{Claude\\Sonnet 5\\(Tesseract input)}},
      {\shortstack{Claude\\Sonnet 5\\(UTRNet input)}}},
    x tick label style={font=\sffamily\scriptsize, align=center},
    axis lines=left, axis line style={-},
    ylabel={CER reduction (\%)},
    font=\sffamily\scriptsize,
    nodes near coords={\pgfmathprintnumber[fixed,fixed zerofill,precision=1,assume math mode=true]{\pgfplotspointmeta}\%},
    every node near coord/.append style={font=\sffamily\scriptsize},
    legend style={at={(0.5,-0.27)}, anchor=north, legend columns=2, draw=none, font=\sffamily\scriptsize, /tikz/every even column/.append style={column sep=10pt}},
    legend cell align=left,
    legend image code/.code={\draw[#1] (0cm,-0.05cm) rectangle (0.35cm,0.05cm);}
]
\addplot[fill=bgray, draw=black!80, line width=0.3pt] coordinates {(1,17.1) (2,-9.0) (3,33.1) (4,3.9) (5,77.6) (6,54.3) (7,64.3) (8,32.6)};
\addplot[fill=bblue, draw=black!80, line width=0.3pt] coordinates {(1,7.4) (2,48.0) (3,0.7) (4,39.0) (5,6.1) (6,17.9) (7,3.1) (8,21.9)};
\legend{Reduction vs. input source CER, Reduction vs. corrector's own Part A CER}
\end{axis}
\end{tikzpicture}
\caption{CER reduction vs.\ the input source and vs.\ each corrector's own Part A CER.}
\label{fig:partb_reframing}
\end{figure*}
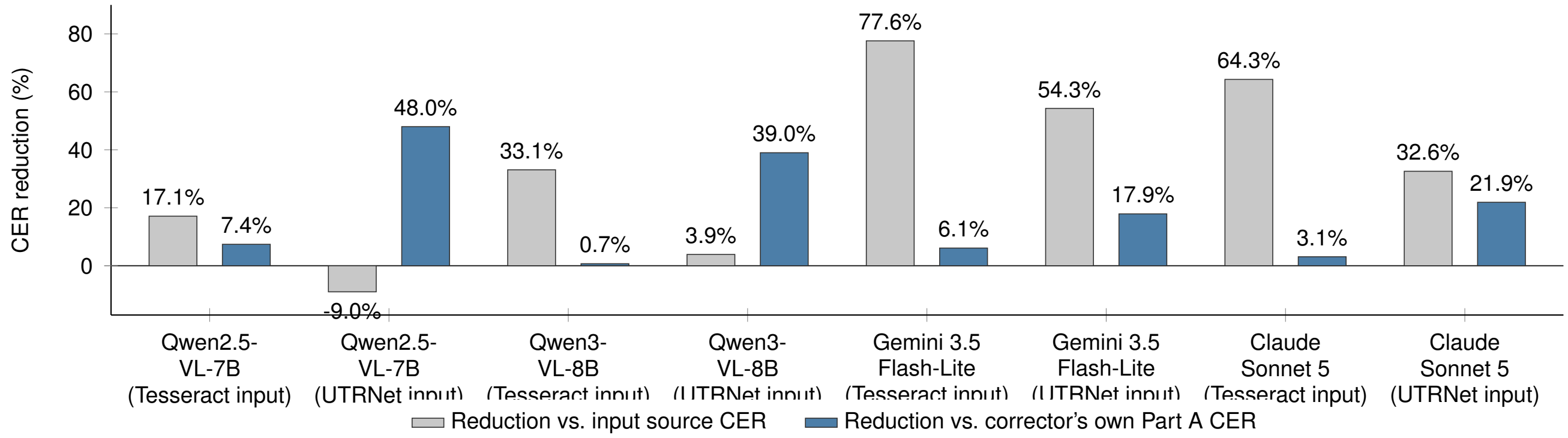

A CER reduction measured against the input source mixes the corrector's own OCR ability with any benefit from the supplied text: Gemini and Sonnet 5 already read the image far better than Tesseract or UTRNet, while the Qwen models read it better than Tesseract but worse than UTRNet. We therefore report two figures for all eight corrector and input combinations (Table~\ref{tab:partb_reduction}, Fig.~\ref{fig:partb_reframing}): the reduction relative to the input source, and the reduction relative to the corrector's own Part A CER on the same images, which isolates whether the supplied text helped. Reductions use unrounded per-image CERs on the images answered in both settings, so they can differ slightly from the rounded table CERs. For Claude Sonnet 5 on Tesseract input the table CER (0.2578, $n=2{,}464$) includes the five images it refused in Part A; on the 2,459 images answered in both settings the CERs are 0.2655 and 0.2574, giving the reported 3.1\%.

\begin{table*}[t]
\caption{Part B results. CER reduction vs.\ the input source and vs.\ the corrector's own Part A CER (PtA), with 95\% bootstrap CIs. Sonnet 5 rows exclude unusable empty outputs ($n=2{,}464$ and $2{,}463$).}
\label{tab:partb_reduction}
\centering
\footnotesize
\begin{tabular*}{\textwidth}{@{}l@{\extracolsep{\fill}}lccc@{}}
\toprule
\textbf{Corrector} & \textbf{Input} & \textbf{CER} & \textbf{Reduction vs. input (\%) [95\% CI]} & \textbf{Reduction vs. own PtA (\%) [95\% CI]} \\
\midrule
Qwen2.5-VL-7B & Tesseract & 0.5984 & 17.1 [15.0, 19.0] & 7.4 [5.0, 9.8] \\
Qwen2.5-VL-7B & UTRNet & 0.3363 & $-9.0$ [$-11.1$, $-7.2$] & 48.0 [46.5, 49.5] \\
Qwen3-VL-8B & Tesseract & 0.4829 & 33.1 [31.3, 34.7] & 0.7 [$-2.1$, $+3.3$] \\
Qwen3-VL-8B & UTRNet & 0.2963 & 3.9 [2.7, 5.2] & 39.0 [37.4, 40.7] \\
Gemini 3.5 Flash-Lite & Tesseract & 0.1613 & 77.6 [76.5, 78.7] & 6.1 [3.9, 8.3] \\
Gemini 3.5 Flash-Lite & UTRNet & 0.1410 & 54.3 [52.4, 56.1] & 17.9 [15.7, 20.2] \\
Claude Sonnet 5 & Tesseract & 0.2578 & 64.3 [63.1, 65.4] & 3.1 [0.4, 6.1] \\
Claude Sonnet 5 & UTRNet & 0.2075 & 32.6 [31.1, 34.2] & 21.9 [19.4, 24.5] \\
\bottomrule
\end{tabular*}
\end{table*}

\subsection{The Same Metric Choice Can Mislead in Either Direction}
\label{sec:crosscorrector}
Table~\ref{tab:partb_reduction} and Fig.~\ref{fig:partb_reframing} show a pattern that neither metric reveals alone. Gemini 3.5 Flash-Lite and Claude Sonnet 5 are strong independent readers in Part A. Given UTRNet's more legible text, both gain by both measures, and all four intervals exclude zero by a wide margin.

For the Qwen models the own-Part-A-relative figures on UTRNet input look even larger (48.0\% and 39.0\%), but this is largely mechanical: UTRNet's text is far better than either model's own zero-shot reading, so adopting it looks like a big gain whether or not correction adds value. The reduction-vs-input column exposes this directly. Given UTRNet's text, Qwen2.5-VL-7B's corrected output is on average worse than UTRNet's raw prediction ($-9.0\%$, 95\% CI $[-11.1\%,-7.2\%]$). Per image, 1,027 of 2,465 (41.7\%) got worse and 997 (40.4\%) improved: an almost even split with a small but statistically confirmed net loss. Qwen3-VL-8B's gain over UTRNet's text is small ($+3.9\%$). We computed per-image counts only for Qwen2.5-VL-7B.

Two results are easy to misread. Qwen3-VL-8B's own-Part-A-relative gain on Tesseract input ($+0.7\%$, CI spanning zero) means correction does not measurably improve on its own zero-shot reading. Claude Sonnet 5's gain on Tesseract input (3.1\%, 95\% CI $[+0.4\%,+6.1\%]$) is the smallest statistically significant effect in the table, and a cluster-aware bootstrap could widen it (Section~\ref{sec:discussion}); Sonnet 5 already reads the image well compared with Tesseract's noisy text, so the text adds little.

No single metric shows whether correction works. Reduction relative to the input is inflated when the corrector reads much better than the input source; reduction relative to the corrector's own Part A score is inflated in the opposite case. Taking both together, only Gemini and Sonnet 5 show statistically distinguishable gains on both input sources: clear on UTRNet's more legible text and marginal on Tesseract's poor output.

\subsection{The Correction Benefit Appears to Depend on Input Legibility}
\label{sec:legibility}
We also asked how the benefit depends on the legibility of the input text. For the four correctors tested on Tesseract output, we binned each image by how legible Tesseract's guess was: decent (CER $\leq$ 0.5), bad ($0.5 <$ CER $\leq$ 0.9), or very poor (CER $>$ 0.9). In each bin we measured each corrector's own-Part-A-relative gain (Table~\ref{tab:legibility_tiers}).

\begin{table*}[t]
\caption{Correction gain by Tesseract input legibility, as percent CER reduction relative to each corrector's own Part A CER (95\% bootstrap CIs).}
\label{tab:legibility_tiers}
\centering
\footnotesize
\begin{tabular*}{\textwidth}{@{}l@{\extracolsep{\fill}}crrrr@{}}
\toprule
\textbf{Tesseract quality} & \textbf{n} & \textbf{Qwen2.5-VL} & \textbf{Qwen3-VL} & \textbf{Gemini} & \textbf{Claude Sonnet 5} \\
\midrule
Decent (CER $\leq$ 0.5) & 266 & +31.7 [27.0, 36.3] & +1.8 [$-9.0$, 11.3] & +14.0 [7.1, 20.4] & +7.7 [0.7, 14.3] \\
Bad (0.5--0.9) & 1849 & +5.9 [3.0, 8.6] & +0.9 [$-2.3$, 4.0] & +6.5 [4.4, 9.3] & +2.6 [$-0.1$, 5.5] \\
Very poor (CER $>$ 0.9) & 350 & $-0.4$ [$-7.4$, 6.0] & $-1.0$ [$-8.5$, 5.8] & +1.8 [$-3.5$, 6.7] & +3.3 [$-5.5$, 14.8] \\
\bottomrule
\end{tabular*}
\end{table*}

The benefit of a noisy OCR prior tends to be larger when the prior is more legible, though not for every corrector. The gain is largest in the decent bin for all four correctors, and for Qwen2.5-VL and Gemini it shrinks as input quality drops. Significant gains appear in the decent bin for Qwen2.5-VL, Gemini and Claude Sonnet 5, in the bad bin only for Qwen2.5-VL and Gemini, and in the very poor bin ($n=350$) for none, so the sign of the very-poor effect for any single corrector should not be over-interpreted. Qwen3-VL shows no significant gain in any bin, and Claude Sonnet 5 is not strictly monotone, though its bad- and very-poor-bin intervals overlap almost completely. Binning by Tesseract's CER also partly bins by image difficulty, so a legibility effect cannot be separated from an easy-image effect with this design.

We did not run this breakdown for UTRNet input (mean CER 0.3084, in the decent range); the larger Gemini and Sonnet 5 gains relative to that input are consistent with the pattern but only suggestive.

\subsection{Catastrophic Failures Were Rarer for the Commercial Correctors}
\label{sec:catastrophic}
In the image-plus-text setting, both open-source correctors sometimes fall into repetition loops. On Tesseract input, loops occurred in 3.37\% of outputs for Qwen2.5-VL-7B, 1.42\% for Qwen3-VL-8B, and 0.00\% for Gemini 3.5 Flash-Lite. Here a loop is an output in which a phrase of two to four words occurs three or more times. Gemini's only failures are one empty output per input source (1 of 2,465 each). Claude Sonnet 5 shows no repetition loops; its failures are empty responses that survived a targeted retry (1 of 2,465 images for Tesseract input and 2 of 2,465 for UTRNet input).

Sonnet 5's unusable empty outputs are excluded from the reported averages, as the five refusals were in Part A. Gemini's two empty outputs and the Qwen repetition-loop outputs are scored as returned, so loops raise the Qwen CERs. Treating the unusable Sonnet or Gemini outputs either way changes a mean CER by under 0.001, so this does not affect any comparison. The shape of the failure differs by model family and rates were lowest for the commercial correctors, but model family, quantization, and serving setup are confounded with corrector strength, so we do not claim a causal trend.

\section{Limitations and Future Work}
\label{sec:discussion}
\textbf{Scope of the comparison.} Results are zero-shot under one English prompt, one run per system, and a single version of each commercial model (the API IDs in Section~\ref{sec:method}). The open-source models ran 4-bit quantized with a 128-token cap, which may understate their full-precision performance. We ran no handwriting-trained recognizer; their published error rates use other datasets or splits, and in-domain they would likely outperform zero-shot VLMs.

\textbf{Data and statistics.} We use one public dataset, so we cannot speak to historical or less controlled handwriting, and UNHD may appear in the VLMs' pretraining data. Six images (036\_2, 036\_3, 036\_4, 029\_002\_8, 077\_002\_8, 096\_002\_8) have ground-truth labels that do not match the handwriting (three truncated, three from a different sentence; UTRNet-Large and Gemini agree with the image); we kept them in all scores. Our bootstrap treats images as independent, though images sharing a sentence or writer may have correlated errors. The closest Part A comparison (Claude Sonnet 5 vs.\ UTRNet-Large) was essentially unchanged under a bootstrap over the 837 sentence templates (same interval to three decimals, $[-0.051,-0.033]$); we ran no cluster-aware bootstrap for Part B or over writer clusters.

\textbf{Part B.} A text-only ablation for Qwen3-VL-8B on Tesseract input gave no improvement over Tesseract's own output (CER 0.7321 vs.\ 0.7215), suggesting the image does the work, but we did not repeat it for the other correctors. The legibility breakdown covers only Tesseract input and cannot separate legibility from image difficulty. The correction prompt also calls the OCR guess low-quality and often wrong, even for UTRNet's more legible text, which may push correctors to over-edit.

\textbf{Future work.} Natural next steps are comparing against handwriting-trained recognizers, testing Urdu-language and neutral correction prompts, repeating the legibility analysis on UTRNet input with cluster-aware intervals, and evaluating on historical or less controlled handwriting.

\section{Conclusion}
Without task-specific fine-tuning, both commercial VLMs achieve lower error than a printed-Urdu recognizer and two open-source VLMs on Urdu Nastaliq handwriting; we did not compare against handwriting-trained recognizers. Aggregate accuracy hides how differently the systems fail: empty output (likely a segmentation mismatch), repetition loops, and explicit refusal on very hard-to-read images, plus confident hallucination in one further open-source VLM (InternVL3.5-8B); a second (MiniCPM-V-4.5) was too unstable to evaluate in full, and both were excluded from the ranking (Section~\ref{sec:negative}).

In the correction study, reporting CER reduction against a weak OCR baseline greatly overstates the benefit of supplying its text. Measured against each corrector's own image-only reading, the gain on Tesseract's text is small, and the smallest (Claude Sonnet 5) is the most sensitive to resampling assumptions. Larger gains that are not mechanical artifacts appear only for the two strongest correctors given UTRNet's more legible text. The benefit appears to depend on the legibility of the supplied text, a weaker corrector can make an already fairly legible text worse, and catastrophic failures were rarer for the commercial correctors.

\end{document}